\documentclass[letterpaper]{article} % DO NOT CHANGE THIS
\usepackage[draft]{aaai25}  % DO NOT CHANGE THIS
\usepackage{times}  % DO NOT CHANGE THIS
\usepackage{helvet}  % DO NOT CHANGE THIS
\usepackage{courier}  % DO NOT CHANGE THIS
\usepackage[hyphens]{url}  % DO NOT CHANGE THIS
\usepackage{graphicx} % DO NOT CHANGE THIS
\usepackage{natbib}  % DO NOT CHANGE THIS AND DO NOT ADD ANY OPTIONS TO IT
\usepackage{caption} % DO NOT CHANGE THIS AND DO NOT ADD ANY OPTIONS TO IT
\usepackage{algorithm}
\usepackage[noend]{algorithmic}
\usepackage{xcolor}

\usepackage{multirow}

\usepackage{newfloat}
\usepackage{listings}
\DeclareCaptionStyle{ruled}{labelfont=normalfont,labelsep=colon,strut=off} % DO NOT CHANGE THIS
\floatstyle{ruled}
\newfloat{listing}{tb}{lst}{}
\floatname{listing}{Listing}
\usepackage{booktabs}
\usepackage{amsmath}
\usepackage{amssymb}
\usepackage{mathtools}
\usepackage{amsthm}
\theoremstyle{plain}
\newtheorem{theorem}{Theorem}[]

\theoremstyle{definition}
\newtheorem{definition}[theorem]{Definition}

\theoremstyle{remark}

\DeclareMathOperator*{\argmax}{arg\,max}

\title{FedLog: Personalized Federated Classification with Less Communication and More Flexibility}

\author {
    Haolin Yu\textsuperscript{\rm 1},
    Guojun Zhang\textsuperscript{\rm 2},
    Pascal Poupart\textsuperscript{\rm 1}
}
\affiliations {
    \textsuperscript{\rm 1}University of Waterloo \& Vector Institute\\
    \textsuperscript{\rm 2}Noah's Ark Lab, Huawei Technologies\\
    h89yu@uwaterloo.ca, guojun.zhang@uwaterloo.ca, ppoupart@uwaterloo.ca
}

\begin{document}

\maketitle

\begin{abstract}
Federated representation learning (FRL) aims to learn personalized federated models with effective feature extraction from local data. FRL algorithms that share the majority of the model parameters face significant challenges with huge communication overhead. This overhead stems from the millions of neural network parameters and slow aggregation progress of the averaging heuristic. To reduce the overhead, we propose to share sufficient data summaries instead of raw model parameters. The data summaries encode minimal sufficient statistics of an exponential family, and Bayesian inference is utilized for global aggregation. It helps to reduce message sizes and communication frequency. To further ensure formal privacy guarantee, we extend it with differential privacy framework. Empirical results demonstrate high learning accuracy with low communication overhead of our method.

%The Bayesian inference step further reduces communication rounds due to its non-iterative nature. Moreover, clients with different computational resources may choose any body architecture flexibly. We further extend it with differential privacy, providing formal guarantees. Our empirical evaluations show we outperform prior arts under various scenarios.
\end{abstract}

% Uncomment the following to link to your code, datasets, an extended version or similar.
%
%\begin{links}
     %\link{Code}{https://aaai.org/example/code}
     %\link{Datasets}{https://aaai.org/example/datasets}
     %\link{Extended version}{https://aaai.org/example/extended-version}
%\end{links}

\section{Introduction}

%As the famous claim states, \textit{applied machine learning is basically feature engineering} \cite{claim}. 

Representation learning plays a crucial role in machine learning by effectively extracting features from raw data to facilitate downstream prediction in various fields \cite{flute}. Combined with federated learning (FL), federated representation learning (FRL) aims to learn personalized federated models and diminish the impact of heterogeneous clients. FRL methods separate a whole neural network into two parts, body and head~\cite{lgfedavg, fedrep, fedper}. The body is a deep network that learns a compact feature representation from the raw data. The head is a shallow network with few layers that make predictions in the representation space. Personalization is achieved by localizing either the body or the head. Parameters of the rest of the model are shared with the server for global aggregation with the FedAvg heuristic \cite{fedavg}. %In this paper, we consider the case that the heads are shared.

FRL utilizing body parameter sharing  faces two significant challenges. \textbf{1) Heavy communication overhead.} Sharing body parameters induces huge communication cost per round due to millions of parameters of the deep learning model~\citep{wen2023survey} since the body contains the majority of the parameters. Furthermore, the averaging heuristic slows down the convergence with heterogeneous clients, resulting in more aggregation rounds. \textbf{2) Rigid model architecture.} Sharing body parameters requires the same body architecture among clients~\citep{fedavg,fedper,fedrep, feddbe, fedbabu, li2021communication}.  However, clients usually possess different amounts of computing resources. Thus, one body architecture may not be suitable for all devices. In this case, clients with limited resources cannot effectively participate. Even though some recent works consider sharing head parameters to reduce communication overhead, their communication efficiency is still limited due to slow averaging \cite{convergence}. %However, most aggregation schemes, including averaging, assume homogeneous architecture among clients 

To tackle the above challenges, our motivation is that a succinct data sharing protocol should minimize the bandwidth usage and does not depend on specific model architecture. Note that sharing model parameters in original FRL can be interpreted as sharing implicit client data summaries, since local model parameters capture the information from the input which can recover the raw data~\cite{mothukuri2021survey}. However, there is no guarantee that the information reflected by these parameters is sufficient or necessary to infer the global model. Instead, we can consider sharing concise, but sufficient client data summaries.

Sharing sufficient data summaries (also known as sufficient statistics \cite{jordan2009chapter8}) of clients offers two benefits. 1) sufficient statistics maintain small data sizes which reduce the bandwidth usage. 2) sufficient statistics are model independent, which allows heterogenous model deployment. Based on this idea, we propose \textbf{Fed}erated Bayesian \textbf{Log}istic Regression (FedLog), a new FRL strategy. We consider the case that the body of the model is localized and the head is updated by the server. FedLog acquires sufficient statistics from the client data encoded by the body with an exponential family distribution. The sufficient statistics from each client are then sent to the server to determine the optimal head parameters by maximizing the posterior with Bayesian inference. The theoretical property of sufficient statistics ensures that sufficient information is captured with fixed size to infer global model parameters. Also, due to the Bayesian inference step, the number of communication rounds is reduced thus improving communication efficiency. Note that since the model body is not shared and summation of sufficient statistics is non-invertible, FedLog also avoids potential privacy attacks through weight manipulation, GAN-based reconstruction, or large model memorization effects \citep{boenisch2021curious,mothukuri2021survey}. However, to ensure a formal privacy guarantee, we further incorporate differential privacy framework to mitigate privacy leakage.

In summary, the paper makes the following contributions:
\begin{itemize}
    \item FedLog: a new FRL algorithm, the model of which is carefully designed to provide a statistical interpretation for representation learning.
    \item Experiments demonstrating FedLog's low communication cost (as small as $0.09\%$ of FedAvg) and fast convergence under multiple scenarios.
    \item Incorporation of DP and demonstration of favorable trade-off between privacy and utility.
\end{itemize}

\section{Related Works}

\subsection{Federated Representation Learning}
\label{pfl}
In FL, there is a collection of clients $c \in S$ wishing to collaborate. Each client holds their own data $\mathcal{D}_c = ({\bf X}_c, {\bf y}_c )$ locally. We seek to train some ML model with parameters $\boldsymbol\theta$ on these client data. Conventionally, we would centralize all the data $\mathcal{D} = \bigcup_{c\in S}\mathcal{D}_c$ and learn the model with $\mathcal D$. This approach becomes infeasible if the clients cannot share their data due to privacy concerns. FL intends to tackle this problem. A trusted central server is allowed to receive and send perturbed matrices that only contain limited information about raw data, such as model parameters. %Most FL algorithms iterate between local optimization and global aggregation. During local optimization at $t^{th}$ iteration, all the clients (or a proportion of the clients in some cases) start with the global consensus parameters $\theta^t$, and run the algorithm $\mathcal{A}$ to update the local models: $\mathcal{M}_{\boldsymbol\theta^{t+1}_c} = \mathcal{A}(\mathcal{D}_c,\boldsymbol\theta^t)$.
%Note the local models could have other local parameters $\Tilde{\boldsymbol{\theta}}$ than $\boldsymbol{\theta}$, but is omitted in the notation for simplicity. Then, clients send potentially compressed model parameters instead of data to the central server. The server aggregates these parameters into the global consensus parameters for the next iteration $\boldsymbol\theta^{t+1}$. 

One challenge of FL is how to aggregate model parameters learnt locally so that the resulting global consensus $\boldsymbol{\theta}^{t+1}$ is a better approximation to the centralization version than the last round $\boldsymbol{\theta}^{t}$ . This is especially challenging when the number of local training epochs $> 1$ \citep{convergence}, due to non-linear loss functions and predictors. Thus, many FL algorithms simply resort to the averaging heuristic \citep{fedavg,lgfedavg,fedper,fedrep,achituve2021personalized}. 

In FL, clients often have different data distribution $\Pr_c({\bf X}), \Pr_c({\bf y})$ or even $\Pr_c({\bf y}|{\bf X})$. This is referred to as heterogeneous or non-i.i.d. clients. The averaging heuristic can drastically harm the global aggregation in terms of convergence rate and model utility with non-i.i.d. clients \cite{li2019convergence}. Thus, personalized FL (PFL) is introduced, where a global model is not mandatory, but each client could have their own model that best fits their data distribution \citep{tan2022towards}. Nevertheless, most algorithms still utilize the averaging heuristic for aggregation.

One line of FRL works approach PFL by localizing either the body (LG-FedAvg \cite{lgfedavg}) or the head (FedPer \cite{fedper}, FedRep \cite{fedrep}) of the client models, and shares the rest with the server for averaging. FedProto \citep{fedproto} localizes the whole model, but averages feature representations by class and forces local models to learn similar representations. FedLog can also be interpreted as a representation learning algorithm, where we learn local representations by all the layers except the last one. However, unlike previous works that share model parameters and heuristically average them, we share sufficient statistics and update the global head with Bayesian inference. 

Another line of works (CCVR \citep{ccvr}, FedPFT \citep{fedpft}) fit Gaussian Mixture Models (GMM) to local features extracted by a globally uniformed body, and shares the natural parameters. The server then draws virtual features from the GMM and trains a global model. It is notable that these algorithms rely on pretrained FedAvg or foundation models to unify the body. Although FedLog also fits an exponential family distribution to local features, we do not need such pretrained models since our bodies are trained locally with any architecture. Also, we work with the canonical parameters and do not need to draw virtual samples from the learnt distribution. 

Other related works include communication efficient FL and Bayesian FL. See Appendix A for more details.

\section{Method}

\subsection{Exponential Family}

We start by introducing the definition of exponential family.
\begin{definition}
\label{def}
Exponential family refers to a set of probability distributions of the following canonical form.
\begin{equation}
    \Pr({\bf x}|\boldsymbol\eta) = h({\bf x})\exp(\boldsymbol{\eta}^{\top}{\bf T}({\bf x})-A(\boldsymbol{\eta}))
\end{equation}
where $h({\bf x}):\mathbb{R}^p \rightarrow \mathbb{R}_{\geq 0}$, ${\bf T}({\bf x}):\mathbb{R}^p \rightarrow \mathbb{R}^d$, $A(\boldsymbol{\eta}): \mathbb{R}^d \rightarrow \mathbb{R}$ are known functions.
\end{definition}
Note that $A(\boldsymbol{\eta})$ is automatically determined by $h({\bf x})$ and ${\bf T}({\bf x})$, since it must normalize the probability density function (p.d.f.), so that the integral of the p.d.f.~equals to $1$: $A(\boldsymbol{\eta}) = \ln{\int_{\bf x}}h({\bf x})\exp(\boldsymbol{\eta}^{\top}T({\bf x}))d{\bf x}$.

Many well-known distributions are included in the exponential family, such as Gaussian, binomial, Poisson, and Bernoulli distribution. We can always transform a distribution represented by natural parameters into its canonical form defined above. For example, a binomial distribution has the following p.d.f.:
$$\Pr(x|p) = {n \choose x}p^x(1-p)^{(n-x)}, x \in \{0, 1, \cdots, n\}$$
It can be rewritten as:
$$\Pr(x|\eta) = {n \choose x}\exp({\eta}x - n\ln({1+e^\eta})), \eta = \ln\frac{p}{1-p}$$

Exponential family has a few desirable properties, which makes it a perfect candidate for information sharing in PFL. 1) The \textbf{sufficient} statistic ${\bf T}({\bf x})$ captures all the information of ${\bf x}$ that can be used to infer $\boldsymbol{\eta}$ (i.e., ${\bf x}{\perp \!\!\! \perp}\boldsymbol{\eta} | {\bf T(x)}$, conditional independence) \citep{jordan2009chapter8}. 2) If ${\bf x}_1, {\bf x}_2, \cdots, {\bf x}_n$ are i.i.d. samples from $\Pr({\bf x}|{\boldsymbol{\eta}})$, the summation of sufficient statistics $\sum_{i=1}^n {\bf T}({\bf x}_i)$ is a \textbf{complete} statistic for $\boldsymbol{\eta}$. It contains only information about $\boldsymbol{\eta}$, without ancillary information \citep{casella2024statistical}. 3) It is the only parametric distribution family with sufficient statistics of \textbf{fixed size} that does not grow with the sample size \citep{koopman1936distributions}. 4) It has known \textbf{conjugate priors} \citep{jordan2009chapter9}, a crucial property for Bayesian inference as we will elaborate later. Formal definitions of these properties are in Appendix B.

\subsection{Model Construction}
We detail our model construction and algorithm with the following notation.
\paragraph{Notation.} Let $\Tilde{\boldsymbol{\theta}}_c$ denote the parameters of a local neural network of client $c$. This network serves as a local body. Let $f_{\Tilde{\boldsymbol{\theta}}_c}: \mathbb{R}^p \rightarrow \mathbb{R}^m$ be the function of the local neural network and $\boldsymbol{\Phi}_c = f_{\Tilde{\boldsymbol{\theta}}_c}({\bf X}_c)$ be the $m$ local features extracted from the local input. For convenience, we designate the first feature to be always $1$. Let $n_{class}$ denote the total number of classes in a classification task, and $n_c$ denote the size of the local dataset. Let ${\bf e}_i \in \mathbb{R}^{n_{class}}$ be the standard basis (i.e. one-hot vector form) of label $i \in \{1, 2,\cdots,n_{class}\}$, and $\otimes: \mathbb{R}^m \times \mathbb{R}^{n_{class}} \rightarrow \mathbb{R}^{m*n_{class}}$ be the Kronecker product.

We assume the joint probability of each data point in $(\boldsymbol{\Phi}_c, {\bf y}_c)$, denoted as $\Pr(\boldsymbol{\phi},y)$, is an exponential family with canonical parameters $\boldsymbol{\eta} \in \mathbb{R}^{m*n_{class}}$. We design the sufficient statistics ${\bf T}(\boldsymbol{\phi}, y)$ and the base measure $h(\boldsymbol{\phi}, y)$ to have the following form.
\begin{align}
    \label{suf}
    {\bf T}(\boldsymbol{\phi}, y) &:= \boldsymbol{\phi}\otimes {\bf e}_y\\
    h(\boldsymbol{\phi}, y) &:= \frac{\exp\left(-\sum_{i=1}^{m} \boldsymbol{\phi}_i^2\right)}{\sqrt{\pi^m}}
    \label{base}
\end{align}
where $\boldsymbol{\phi}_i \in \mathbb{R}$ denotes the $i^{th}$ entry of $\boldsymbol{\phi}$.
One advantage with this specific choice of $\bf T$ and $h$ lies in the resulting conditional likelihood $\Pr(y|\boldsymbol{\phi},\boldsymbol{\eta})$. Let $\boldsymbol{\eta}_{y} \in \mathbb{R}^{m}$ denote the $((y-1)*m)^{th}$ to $(y*m)^{th}$ entries of $\boldsymbol{\eta}$, $y \in \{1,2,\cdots,n_{class}\}$. Then,
\begin{align}
    \label{likeli}
    %\Pr(\boldsymbol{\phi},y|\boldsymbol{\eta}) &= \frac{\exp(\boldsymbol{\eta}_y^\top\boldsymbol{\phi}-\boldsymbol{\phi}^\top\boldsymbol{\phi})}{\sqrt{\pi^m}\sum_{y'=1}^{n_{class}}\exp(\boldsymbol{\eta}_{y'}^\top\boldsymbol{\eta}_{y'}/4)}\\
    \Pr(\boldsymbol{\phi},y|\boldsymbol{\eta}) &= \frac{\exp(\boldsymbol{\eta}_y^\top\boldsymbol{\phi}-\boldsymbol{\phi}^\top\boldsymbol{\phi}-A({\boldsymbol{\eta}}))}{\sqrt{\pi^m}}\\
    \Pr(y|\boldsymbol{\phi},\boldsymbol{\eta})&= \frac{\Pr(\boldsymbol{\phi}, y|\boldsymbol{\eta})}{\Pr(\boldsymbol{\phi}|\boldsymbol{\eta})} = \frac{\Pr(\boldsymbol{\phi}, y|\boldsymbol{\eta})}{\sum_{y'=1}^{n_{class}}\Pr(\boldsymbol{\phi}, y'|\boldsymbol{\eta})}\\
    \label{cond}
    &=\frac{\exp(\boldsymbol{\eta}_y^\top\boldsymbol{\phi})}{\sum_{y'=1}^{n_{class}}\exp(\boldsymbol{\eta}_{y'}^\top\boldsymbol{\phi})}
\end{align}
Eq. \ref{cond} is exactly the softmax function over $\boldsymbol{\eta}_y^\top\boldsymbol{\phi}$. This means we can take any deep neural network that extracts $m$ features, and append $\boldsymbol{\eta}$ as the last linear layer that maps the features to $n_{class}$ logits. Then, this composed neural network serves as a stand-alone classifier that computes the conditional likelihood $\Pr(y|{\boldsymbol{\phi}},\boldsymbol{\eta})$. However, Eq. \ref{cond} cannot be directly maximized at the server, since it requires knowledge of uncompressed representation-label pairs. Instead, we utilize Bayesian inference to optimize $\boldsymbol{\eta}$.

With Bayesian inference, $\boldsymbol{\eta}$ is treated as a random variable. A prior distribution $\Pr(\boldsymbol\eta)$ can be specified to incorporate prior knowledge. Then, by Bayes' Theorem, the posterior distribution is:
\begin{align}
     \Pr(\boldsymbol\eta|\boldsymbol{\phi},y) &= \frac{\Pr(\boldsymbol{\phi},y|\boldsymbol\eta)\Pr(\boldsymbol\eta)}{\Pr(\boldsymbol{\phi},y)} \\
     &=  \frac{\Pr(\boldsymbol{\phi},y|\boldsymbol\eta)\Pr(\boldsymbol\eta)}{\sum_{y'}\int_{\boldsymbol{\eta}}\Pr(\boldsymbol{\phi},y'|\boldsymbol\eta)\Pr(\boldsymbol\eta)d\boldsymbol\eta}
     \label{bayes}
\end{align}
The integral in Eq.~\ref{bayes} and thus the posterior $\Pr(\boldsymbol\eta|\boldsymbol{\phi},y)$ may not be tractable for arbitrary priors $\Pr(\boldsymbol\eta)$. A convenient choice that guarantees analytical solutions is the conjugate prior. Given a likelihood $\Pr(\boldsymbol{\phi},y|\boldsymbol\eta)$, a prior is called its conjugate prior if $\Pr(\boldsymbol\eta)$ and $\Pr(\boldsymbol{\phi},y|\boldsymbol\eta)$ follow the same distribution family. Specifically, if the likelihood is an exponential family, it has known conjugate priors \citep{jordan2009chapter9}.
\begin{align}
    \text{Prior:} & \Pr(\boldsymbol{\eta};\boldsymbol{\chi},\nu) = f(\boldsymbol{\chi}, \nu)\exp(\boldsymbol{\eta}^{\top}\boldsymbol{\chi}-{\nu}A(\boldsymbol{\eta}))\\
    \text{Posterior:} & \Pr(\boldsymbol{\eta}|\boldsymbol{\phi},y) = \Pr(\boldsymbol{\eta}; \boldsymbol{\chi}+{\bf T}(\boldsymbol{\phi},y), \nu+1)
    \label{post}
\end{align}
where $\boldsymbol{\chi}\in \mathbb{R}^d, \nu \in \mathbb{R}$ are deterministic parameters of the prior, and $f(\boldsymbol{\chi}, \nu): \mathbb{R}^d \times \mathbb{R} \rightarrow \mathbb{R}$ is automatically determined by $A(\boldsymbol{\eta})$: $f(\boldsymbol{\chi}, \nu)^{-1} = \int_{\boldsymbol{\eta}}\exp(\boldsymbol{\eta^\top\boldsymbol{\chi}}-{\nu}A(\boldsymbol{\eta}))d\boldsymbol{\eta}$.

Another advantage of our model design is that $A$ has an explicit expression.  Let $\boldsymbol{\eta}_{y,i} \in \mathbb{R}$ denote the $i^{th}$ entry of $\boldsymbol{\eta}_{y}$.
\begin{align}
\begin{split}
    %A(\boldsymbol{\eta}) &= \ln\sum_{y=1}^{n_{class}}\int_{-\infty}^{\infty}h(\boldsymbol{\phi}, y)\exp(\boldsymbol{\eta}_y^{\top}{\bf T}({\boldsymbol{\phi}}, y))d{\boldsymbol{\phi}}\\
    A(\boldsymbol{\eta}) &= \ln\sum_{y=1}^{n_{class}}\int_{-\infty}^{\infty}\frac{\exp\left(\sum_{i=1}^m\boldsymbol{\eta}_{y,i}\boldsymbol{\phi}_i-\boldsymbol{\phi}_i^2\right)}{\sqrt{\pi^m}}d{\boldsymbol{\phi}} \\
    &= \ln\sum_{y=1}^{n_{class}}\exp\left(\frac{\sum_{i=1}^m\boldsymbol{\eta}_{y,i}^2}{4}\right) \label{eq:closed-form}
    %&= \ln\sum_{y=1}^{n_{class}}\exp\left(\frac{\boldsymbol{\eta}_y^\top\boldsymbol{\eta}_y}{4}\right)
\end{split}
\end{align}
Due to this analytical solution, we can directly optimize Eq.~\ref{post} without further approximations.
%Unfortunately, the normalizing constant function $f(\boldsymbol{\chi},\nu)$, the marginal likelihood $\Pr(y|{\boldsymbol{\phi}})$, and the predictive posterior $\Pr(y_*|\boldsymbol{\phi}_*,\boldsymbol{\phi},y)$ are still intractable.

\subsection{FedLog}

Based on the above model, we propose our new algorithm FedLog, summarized in Algo.~\ref{alg:client}. At the beginning, the server initializes $\boldsymbol{\eta}$ (the global head) randomly. The clients initialize $\Tilde{\boldsymbol{\theta}}_c$ (the local bodies) either completely randomly, or with the same random seed sent by the server to unify the initialization. Note $\Tilde{\boldsymbol{\theta}}_c$ is not part of our exponential family assumption, thus we do not require them to have the same shape or architecture amongst different clients.

Parameters that we need to optimize are essentially $\Tilde{\boldsymbol{\theta}}_c$ and $\boldsymbol{\eta}$, which can be done by maximizing $\Pr(y|\boldsymbol{\phi},\boldsymbol{\eta})$ and $\Pr(\boldsymbol{\eta}|\boldsymbol{\phi}, {y})$ in turns iteratively, similarly to the expectation-maximization algorithm. Concretely, all the clients $c \in S$ first fix the global head $\boldsymbol{\eta}$, and update their local bodies $\Tilde{\boldsymbol{\theta}}_c$ with gradient descent. We derive the loss function as:
\begin{align}
\begin{split}
    %\mathcal{L}_c &= -\ln\Pr({\bf y}_c|\boldsymbol{\Phi}_c, \boldsymbol{\eta})\\
    \mathcal{L}_c &= -\sum_{i=1}^{n_c}\ln\Pr({\bf y}_{c,i}|\boldsymbol{\Phi}_{c,i}, \boldsymbol{\eta}) \\
    &= -\sum_{i=1}^{n_c}\ln\frac{\exp(\boldsymbol{\eta}_{y_{c,i}}^\top\boldsymbol{\Phi}_{c,i})}{\sum_{y=1}^{n_{class}}\exp(\boldsymbol{\eta}_{y}^\top\boldsymbol{\Phi}_{c,i})}
    \label{softmax}
\end{split}
\end{align}
where $(\boldsymbol{\Phi}_{c,i}, {\bf y}_{c,i}) \in \mathbb{R}^m \times \mathbb{R}$ is the $i^{th}$ data point of $\boldsymbol{\Phi}_{c}, {\bf y}_{c}$. This is exactly the cross entropy loss widely used in deep learning for classification tasks. Then, clients compute sufficient statistics of their local data $\sum_{i=1}^{n_c}{\bf T}(\boldsymbol{\Phi}_{c,i}, {\bf y}_{c,i}) = \sum_{i=1}^{n_c}\boldsymbol{\Phi}_{c,i}\otimes{\bf e}_{{\bf y}_{c,i}}$. They then send the summations and $n_c$ to the server for global head learning. As we have discussed with the introduction of exponential family, the sufficient statistics contain all information in the representations that could be used to infer $\boldsymbol{\eta}$ in our model. The server only needs to know the summation of all the sufficient statistics, since
\begin{align}
\begin{split}
    & \Pr(\boldsymbol{\eta}|\boldsymbol{\Phi}_{c_1}, {\bf y}_{c_1},\cdots,\boldsymbol{\Phi}_{c_k}, {\bf y}_{c_k}) \\
    &= \Pr(\boldsymbol{\eta}; \boldsymbol{\chi}+\sum_{c\in S}\sum_{i=1}^{n_c}\boldsymbol{\Phi}_{c,i}\otimes{\bf e}_{{\bf y}_{c,i}}, \nu+\sum_{c\in S}n_c)
\end{split}
\end{align}
can be trivially inferred from Eq.~\ref{post}. Note the size of the message sent by each client equals the size of the last linear layer. After receiving the sufficient statistics, the server  computes $\boldsymbol{\Phi} = \sum_{c \in S}\sum_{i=1}^{n_c}\boldsymbol{\Phi}_{c,i}\otimes{\bf e}_{{\bf y}_{c,i}}, n = \sum_{c \in S}n_c$ and updates the global head $\boldsymbol{\eta}$ by maximum a posteriori (MAP):
\begin{align}
\begin{split}
    \boldsymbol{\eta} &= \argmax_{\boldsymbol{\eta}} \ln\Pr(\boldsymbol{\eta};\boldsymbol{\chi}+\boldsymbol{\Phi}, \nu+n) \\
    &= \argmax_{\boldsymbol{\eta}} \ln\frac{\exp\left(\boldsymbol{\eta}^\top(\boldsymbol{\chi}+\boldsymbol{\Phi})\right)}{(\sum_{y=1}^{n_{class}}\exp(\boldsymbol{\eta}_y^\top\boldsymbol{\eta}_y/4))^{(\nu+n)}}
\label{kernel}
\end{split}
\end{align}

\begin{algorithm}[tb]
   \caption{FedLog (${\bf X}_c, {\bf y}_c$: local data, $\Tilde{\boldsymbol{\theta}}_c$ local body parameters, $\boldsymbol{\eta}$: global head parameters, $\boldsymbol{\chi}$: prior parameter, $\nu$: prior parameter, $\zeta$: local learning rate)}
   \label{alg:client}
\begin{algorithmic}
    \STATE {\bfseries Server:} initializes $\boldsymbol{\eta}$
    \FOR{each {\bfseries client} $c \in S$}
        \STATE Initialize $\Tilde{\boldsymbol{\theta}}_c$
    \ENDFOR
    \FOR{each global update round}
    \STATE {\bfseries Server:} sends $\boldsymbol{\eta}$ to clients
    \FOR{each {\bfseries client} $c \in S$}
    \FOR{each local update round}
    \STATE $\Tilde{\boldsymbol{\theta}}_c \gets \Tilde{\boldsymbol{\theta}}_c - \zeta\nabla\mathcal{L}_c$ (Eq.~\ref{softmax})
    \ENDFOR
    \STATE $\boldsymbol{\Phi}_{c} \gets f_{\Tilde{\boldsymbol{\theta}}_c}({\bf X}_c)$
    \STATE Send $\sum_{i=1}^{n_c}\boldsymbol{\Phi}_{c,i}\otimes{\bf e}_{{\bf y}_{c,i}}, n_c$ to server
    \ENDFOR
    \STATE {\bfseries Server:}
    \STATE $\boldsymbol{\Phi} \gets \sum_{c \in S}\sum_{i=1}^{n_c}\boldsymbol{\Phi}_{c,i}\otimes{\bf e}_{{\bf y}_{c,i}}, n \gets \sum_{c \in S}n_c$
    \STATE $\boldsymbol{\eta} \gets \argmax_{\boldsymbol{\eta}} \Pr(\boldsymbol{\eta};\boldsymbol{\chi}+\boldsymbol{\Phi}, \nu+n)$ (Eq. ~\ref{kernel})
    \ENDFOR
\end{algorithmic}
\end{algorithm}

Note Eq.~\ref{kernel} is a convex optimization task. We can easily compute its sole maximum by gradient descent with  complexity $O(m*n_{class})$. Then, the server sends $\boldsymbol{\eta}$ back to all the clients and starts the next round of updates. The process is repeated until convergence.

\begin{figure*}
\includegraphics[width=5.8cm,height=3.3cm]{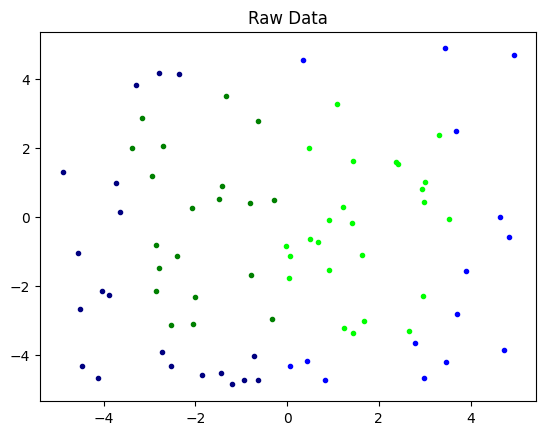}
\includegraphics[width=5.8cm,height=3.3cm]{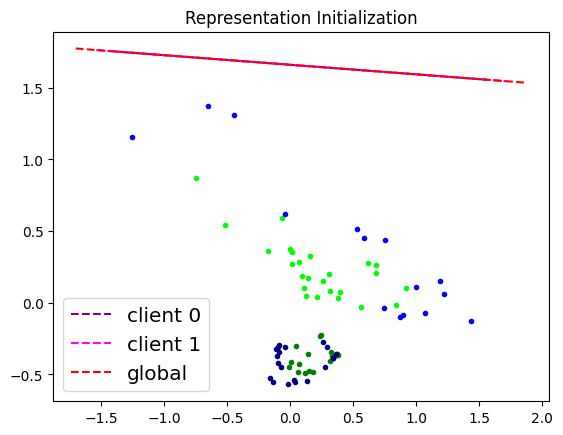}
\includegraphics[width=5.8cm,height=3.3cm]{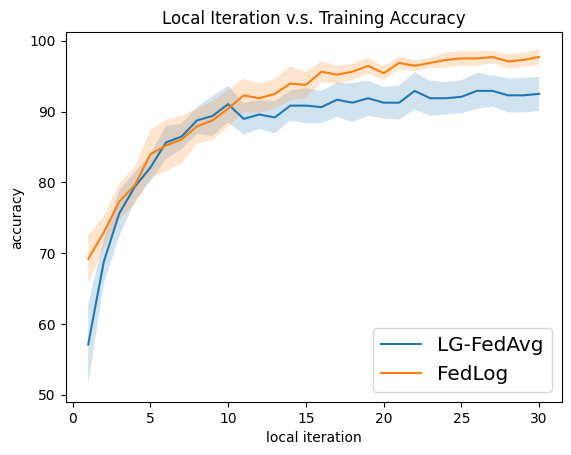} \\
\includegraphics[width=5.8cm,height=3.3cm]{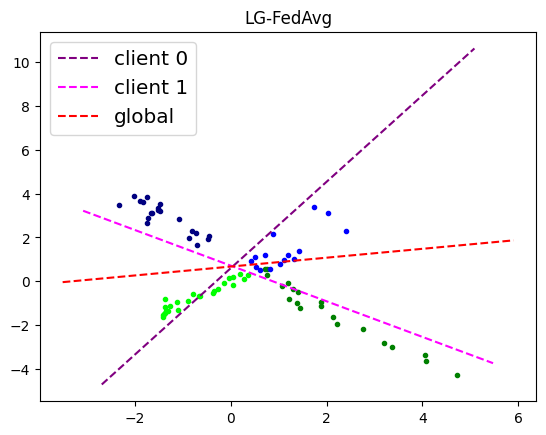}    \includegraphics[width=5.8cm,height=3.3cm]{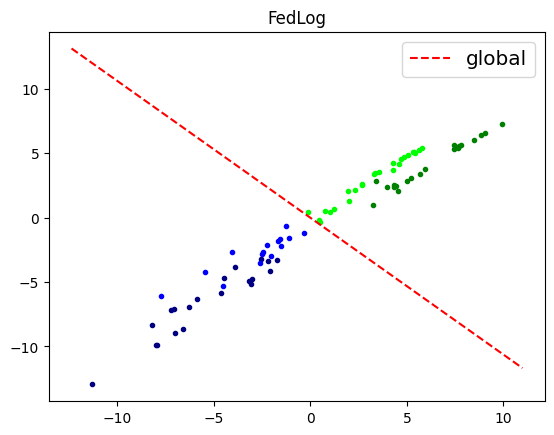}    \includegraphics[width=5.8cm,height=3.3cm]{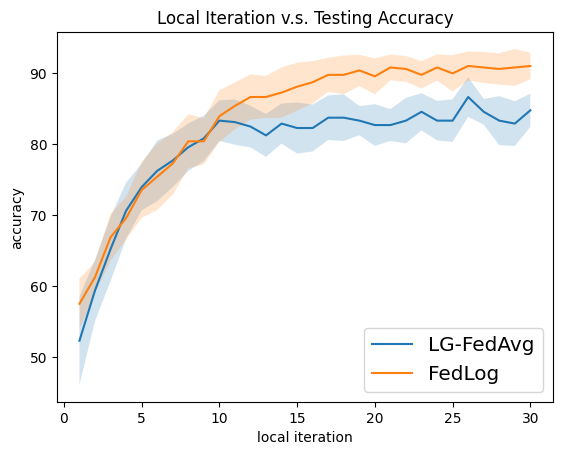}
    \caption{Synthetic experiments. Dots are data points or local representations: dark green: client 0 class 0; light green: client 1 class 0; Dark blue: client 0 class 1; light blue: client 1 class 1. Dashed lines are linear separators. Accuracy results are averaged over 6 seeds.} %Subfigures from top to bottom, left to right shows: raw data; initial local representation; training accuracy; local representation and linear separator learnt by LG-FedAvg; local representation and linear separator learnt by FedLog; testing accuracy.}
    \label{fig:syn}
\end{figure*}
\subsection{Interpretation and FedLog-C}

In this section, we analyze our assumptions in more details and give some insights about how FedLog works. The first assumption we made is that $\forall c \in S$, the local data points $\boldsymbol{\Phi}_c, {\bf y}_c$ are from the same exponential family distribution whose pdf is given by Eq.~\ref{likeli}. In other words, we assume that the local bodies $\Tilde{\boldsymbol{\theta}}_c$ transform their input, of any form, to the same representation space. This may seem infeasible at first glance since we do not directly aggregate the local body parameters, and they may follow any architecture. However, note we fix the global head $\boldsymbol{\eta}$ during local updates, by which the local bodies are forced to learn a universal representation space. See the synthetic experiment below for more details. This assumption allows principled discriminative training for the local bodies with cross entropy loss, but unavoidably leaves a generative model for learning the global head. We can further see that $\Pr(\boldsymbol{\phi}|y, \boldsymbol{\eta}) \propto \exp(\boldsymbol{\eta}_y^\top\boldsymbol{\phi}-\boldsymbol{\phi}^\top\boldsymbol{\phi})$, which is the kernel of a multivariate Gaussian distribution. This means we essentially assumed a mixture of Gaussians for the local features $\boldsymbol{\phi}$. To mitigate the gap between the assumption and the actual feature distribution, we propose a variation FedLog-C. An auxiliary loss is added during local training to force the local bodies to learn Gaussian-like clusters. Let $\boldsymbol{\Phi}_y$ denote the $((y-1)*m)^{th}$ to $(y*m)^{th}$ entries of $\boldsymbol{\Phi}$. Let $\boldsymbol{\Phi}_{y,0}$ denote the first entry of $\boldsymbol{\Phi}_y$. Then $\overline{\boldsymbol{\Phi}_y} = \boldsymbol{\Phi}_y / \boldsymbol{\Phi}_{y,0}$ is the global mean representation of class $y$. Inspired by contrastive learning \cite{schroff2015facenet}, we derive the new local loss as:
\begin{align}
\begin{split}
\label{fedlogC}
    \mathcal{L}'_c = \mathcal{L}_c &+ \alpha \sum_{i=1}^{n_c}(\boldsymbol{\Phi}_{c,i}\otimes{\bf e}_{{\bf y}_{c,i}}-\overline{\boldsymbol{\Phi}_{{\bf y}_{c,i}}})^2/n_c\\
    &- \beta \sum_{y' \neq {\bf y}_{c,i}}\sum_{i=1}^{n_c}(\boldsymbol{\Phi}_{c,i}\otimes{\bf e}_{{\bf y}_{c,i}}-\overline{\boldsymbol{\Phi}_{y'}})^2/n_c
\end{split}
\end{align}
$\alpha \in \mathbb{R}_{\geq 0}$ controls how compact the clusters should be. $\beta \in \mathbb{R}_{\geq 0}$ controls the distance between different clusters. Since clients need to know $\boldsymbol{\Phi}$, the server simply broadcasts the aggregated statistic to all the clients. Clients can optimize the same $\boldsymbol{\eta}$ locally, preserving the same communication cost. %This variation is summarized in Alg.~ \ref{alg:C}. 

The second assumption we made is the prior of the global head. Since $\boldsymbol{\eta}$ has a support over $\mathbb{R}^{m*n_{class}}$, it is impossible to specify a uniform prior. Without any prior knowledge, we can set ${\boldsymbol{\chi}} = {\bf 0}, \nu = 1$. The prior then becomes $\Pr(\boldsymbol{\eta}) \propto \exp(-A(\boldsymbol{\eta})) = (\sum_{y=1}^{n_{class}}\exp(\boldsymbol{\eta}_y^\top\boldsymbol{\eta}_y/4))^{-1}$. The p.d.f. takes its maximum at $\boldsymbol{\eta} = {\bf 0}$ and decreases quickly as the absolute value of entries of $\boldsymbol{\eta}_y$ grows larger. This is in analogy to the Lasso regularizer in the regression case, which prevents the model from learning coefficients with large absolute values due to noise or over-fitting.

From the Bayesian view, FedLog can take any deep classifier, and make the last linear layer Bayesian. It essentially operates a Bayesian logistic regression model on the local representations. We start from a generative assumption and achieve the cross-entropy loss conditional likelihood usually assumed directly in Bayesian logistic regression. We obtained an analytical solution for the kernel of the posterior, which can be calculated easily by the summation of sufficient statistics. The shared statistic cannot be further compressed without losing information from the representations. We formalize this statement with the following theorem.
\begin{theorem}
    If $({\boldsymbol{\phi}_1}, y_1), ({\boldsymbol{\phi}_2}, y_2), \cdots, ({\boldsymbol{\phi}_n}, y_n)$ are i.i.d. samples from the exponential family defined with Eq. \ref{likeli}, then ${\bf T}(({\boldsymbol{\phi}_1}, y_1), \cdots, ({\boldsymbol{\phi}_n}, y_n)) = \sum_{i=1}^n {\boldsymbol{\phi}_i}\otimes{\bf e}_{y_i}$ is a minimal sufficient statistic independent of every ancillary statistic. \label{thm:sufficient-statistics}
\end{theorem}
\begin{proof}
    See Appendix B for the proof.
\end{proof}
\iffalse
\begin{proof}
    ${\bf T}(({\boldsymbol{\phi}_1}, y_1), ({\boldsymbol{\phi}_2}, y_2), \cdots,({\boldsymbol{\phi}_n}, y_n))$ is trivially sufficient and complete since $({\boldsymbol{\phi}_i}, y_i)$ follows an exponential family distribution. Define
\begin{align*}
    R &= \frac{\Pr(({\boldsymbol{\phi}_1}, y_1), ({\boldsymbol{\phi}_2}, y_2), \cdots, ({\boldsymbol{\phi}_n}, y_n)|\boldsymbol{\eta})}{\Pr(({\boldsymbol{\phi}'_1}, y'_1), ({\boldsymbol{\phi}'_2}, y’_2), \cdots, ({\boldsymbol{\phi}'_n}, y'_n)|\boldsymbol{\eta})} \\
    &= \exp(\boldsymbol{\eta}^\top(\sum_{i=1}^n {\boldsymbol{\phi}_i}\otimes{\bf e}_{y_i}-\sum_{i=1}^n {\boldsymbol{\phi}'_i}\otimes{\bf e}_{y'_i}))
\end{align*}
${\bf T}$ is also minimal since $R$ is independent of $\boldsymbol{\eta}$ if and only if $${\bf T}(({\boldsymbol{\phi}_1}, y_1), \cdots, ({\boldsymbol{\phi}_n}, y_n)) = {\bf T}(({\boldsymbol{\phi}'_1}, y'_1), \cdots ({\boldsymbol{\phi}'_n}, y'_n))$$
See Appendix B for further details.
\end{proof}
\fi
From the federated representation learning view, FedLog has a one-layer global head and a deep local body. It iterates between learning local representations and learning global linear separators as if it has seen all the local representations. The two learning processes are completely separated, unlike the common paradigm where the local representations and the linear separators are often optimized jointly. The clients are only responsible for moving local representations to the correct sides of the fixed linear separator. The server is only responsible for finding the best linear separator given the local representations. As we will show with the experiments, we can converge faster than using the averaging heuristic.

\subsection{Differential Privacy}
FL can be combined with formal mechanisms such as differential privacy (DP) \citep{wei2020federated,triastcyn2019federated} or secure multi-party computation (MPC)~\citep{truex2019hybrid,byrd2020differentially,li2020privacy}, to provide formal privacy guarantees. We now extend FedLog to be differentially private.

($\epsilon, \delta$)-DP protects clients' privacy by adding noise to the shared information so that the adversaries cannot effectively tell if any record is included in the dataset (controlled by $\epsilon > 0$) at most times (controlled by $0 \leq \delta < 1$) \citep{dp}.
\begin{definition}
    A mechanism $M_{DP}$ satisfies $(\epsilon$, $\delta)$-DP if for any two datasets $\mathcal{D}, \mathcal{D}'$ that differ by only one record (i.e. $|(\mathcal{D}-\mathcal{D}') \cup (\mathcal{D}'-\mathcal{D})|=1$), and for any possible output $O \in Range(M_{DP})$,
    $$\Pr_{O \sim M_{DP}(\mathcal{D})}\left[\log(\frac{\Pr[M_{DP}(\mathcal{D})=O]}{\Pr[M_{DP}(\mathcal{D}')=O]}) > \epsilon \right] < \delta$$
\end{definition}
Intuitively, $(\epsilon, \delta)$-DP guarantees that the inner log ratio, considered as the information loss leaked to the adversaries, is bounded by the privacy budget $\epsilon$ with probability $\delta$. We add Gaussian noise to shared sufficient statistics as follows.
\begin{theorem}
    If the absolute value of features are clipped to $b$ and there are in total $k$ global update rounds, FedLog messages satisfy $(\epsilon, \delta)$-DP with additive Gaussian noise ${\bf T}'(\boldsymbol{\Phi}_c, {\bf y}_c) := {\bf T}(\boldsymbol{\Phi}_c, {\bf y}_c) +\mathcal{N}({\bf 0}, \sigma^2{\bf I})$, where $\sigma = \sqrt{8k(1+(m-1)*b^2)\ln(e+\epsilon/\delta)}/{\epsilon}$.
\end{theorem}
\begin{proof}
\begin{align*}
    \max_{\mathcal{D},\mathcal{D}'}||{\bf T}(\mathcal{D})-{\bf T}(\mathcal{D}')||_2 
    &= \max_{\boldsymbol{\phi}, y}||{\bf T}(\boldsymbol{\phi}, y)||_2 \\ &= \sqrt{1+(m-1)*b^2}
\end{align*}
    See more details in Appendix C.
\end{proof}

When $n$ is large enough, the amount of noise (independent of $n$) becomes negligible to the model. Optionally, with secure MPC or central DP, it is sufficient to add such Gaussian noise once globally to $\boldsymbol{\Phi}$ each global update round. The server then computes the posterior normally with the noisy sufficient statistics. 

{
\begin{table*}
  \caption{Testing accuracy and communication cost reported for MNIST, CIFAR10, and CIFAR100. Accuracy reports the mean $\pm$ standard error of the testing accuracy over $10$ seeds. Higher is better. Communication cost reports the total message size transmitted between clients and the server. Lower is better. $\Uparrow$ denotes significantly higher results with $p < 0.01$; $\Downarrow$ denotes significantly lower results with $p < 0.01$.}
  \centering
 {
 \small
  \begin{tabular}{ccccccc}
    \toprule
    %& FedAvg & LG-FedAvg 1 & LG-FedAvg 2 & FedPer & FedRep & CS-FL & {\bf FedLog (ours)}\\
    & \multicolumn{2}{c}{MNIST} & \multicolumn{2}{c}{CIFAR10} & \multicolumn{2}{c}{CIFAR100}\\
    \cmidrule(r){2-3}
    \cmidrule(r){4-5}
    \cmidrule(r){6-7}
    & accuracy & comm cost & accuracy & comm cost & accuracy & comm cost\\
    \midrule
    FedAvg & 89.76$\pm$0.69$^\Downarrow$ & 3.45$\pm$0.02Gb$^\Uparrow$ & 26.29$\pm$0.44$^\Downarrow$ & 37.9$\pm$1.31Gb$^\Uparrow$ & 13.34$\pm$0.15$^\Downarrow$ & 69.1$\pm$0.47Gb$^\Uparrow$\\
    LG-FedAvg 1 & 97.85$\pm$0.05$^\Downarrow$ & 4.81$\pm$0.31Mb$^\Uparrow$ & 86.57$\pm$0.29$^\Downarrow$ & 0.26$\pm$0.02Gb$^\Uparrow$ & 55.00$\pm$0.26$^\Downarrow$ &  4.34$\pm$0.18Gb$^\Uparrow$\\
    LG-FedAvg 2 & {98.18$\pm$0.06$^\Downarrow$} & 0.16$\pm$0.07Gb$^\Uparrow$ & 85.56$\pm$0.32$^\Downarrow$ & 3.38$\pm$0.32Gb$^\Uparrow$ & 54.90$\pm$0.24$^\Downarrow$ &  9.53$\pm$0.53Gb$^\Uparrow$\\
    FedPer & 96.16$\pm$0.19$^\Downarrow$ & 0.65$\pm$0.05Gb$^\Uparrow$ & 83.54$\pm$0.40$^\Downarrow$ & 27.7$\pm$1.50Gb$^\Uparrow$ & 52.82$\pm$0.21$^\Downarrow$ & 42.7$\pm$1.47Gb$^\Uparrow$\\
    FedRep & 95.51$\pm$0.29$^\Downarrow$ & 36.3$\pm$3.38Mb$^\Uparrow$ & 82.96$\pm$0.35$^\Downarrow$ & 15.3$\pm$1.33Gb$^\Uparrow$ & 48.70$\pm$0.29$^\Downarrow$ & 11.0$\pm$0.40Gb$^\Uparrow$\\
    CS-FL & 79.65$\pm$1.22$^\Downarrow$ & 0.35$\pm$0.01Gb$^\Uparrow$ & 23.60$\pm$1.08$^\Downarrow$ & 2.72$\pm$0.41Gb$^\Uparrow$ & 4.52$\pm$0.15$^\Downarrow$ & 13.7$\pm$0.22Gb$^\Uparrow$ \\
    FedBabu & 86.30$\pm$1.04$^\Downarrow$ & 2.25$\pm$0.02Gb$^\Uparrow$ & 25.37$\pm$0.44$^\Downarrow$ & 34.7$\pm$2.29Gb$^\Uparrow$ & 9.70$\pm$0.16$^\Downarrow$ & 59.3$\pm$0.31Gb$^\Uparrow$ \\
    FedProto & {98.19$\pm$0.06$^\Downarrow$} & {\bf 3.02$\pm$0.17Mb} & {87.37$\pm$0.26}$^\Downarrow$ & 0.18$\pm$0.01Gb$^\Uparrow$ & 55.32$\pm$0.19$^\Downarrow$ & 3.01$\pm$0.12Gb$^\Uparrow$\\
    FedDBE & 96.79$\pm$0.34$^\Downarrow$ & 1.71$\pm$0.39Gb$^\Uparrow$ & 72.77$\pm$0.79$^\Downarrow$ & 38.3$\pm$0.95Gb$^\Uparrow$ & 36.67$\pm$0.85$^\Downarrow$ & 55.1$\pm$2.80Gb$^\Uparrow$ \\
    \midrule
    {\bf FedLog (ours)} & {98.15$\pm$0.05$^\Downarrow$} & {\bf 3.18$\pm$0.31Mb} & {87.08$\pm$0.22$^\Downarrow$} & { 0.14$\pm$0.01Gb$^\Uparrow$} & { 56.46$\pm$0.27$^\Downarrow$} & {\bf 2.38$\pm$0.09Gb$^\Downarrow$} \\
    {\bf FedLog-C (ours)} & {\bf 98.41$\pm$0.07} & {\bf 3.18$\pm$0.15Mb} & {\bf 87.57$\pm$0.25} & {\bf 0.11$\pm$0.01Gb} & {\bf 56.78$\pm$0.26} & 2.74$\pm$0.12Gb\\
    \bottomrule
  \end{tabular}
  }
\label{Tab:comm}
\end{table*}
}
\section{Experiments}
\subsection{Synthetic}
\label{syn}

We designed the following synthetic experiment to justify our claim that FedLog can learn universal local representation spaces without sharing local feature extractors, and argue why FedLog can converge faster than prior arts. As shown in the top left image of Fig.~\ref{fig:syn}, we first sample $80$ two-dimensional training data points $(x_1, x_2)$ uniformly from the $[-5,5]\times[-5,5]$ square. Data points are separated into class 0 (blue dots) and class 1 (green dots) by the circle located at the origin and of radius $26/7$. These data points are further divided evenly into two sets, client 0 (dark dots) and client 1 (light dots), based on ordered $x_1$ to simulate non-i.i.d. clients. Client 0 operates a three-layer fully connected feature extractor, while client 1 operates a two-layer fully connected feature extractor, to simulate clients with different computational resources and showcase the flexibility. We compare FedLog to LG-FedAvg with a one-layer global head, in which case the size of shared messages is the same between those two algorithms. The top middle image of Fig.~\ref{fig:syn} shows the local representations and linear separators with random initialization. Dashed lines are the linear separators induced by the global head. To be fair, the head of both clients are initialized to be the same for LG-FedAvg. We run FedLog and LG-FedAvg for one global update round, with $1$ to $30$ local iterations. The bottom left and middle images of Fig.~\ref{fig:syn} respectively show the models LG-FedAvg and FedLog converge to locally. LG-FedAvg learns very different local representations and linear separators even if the last layer is initialized to be the same, and the averaged global linear separator (red dashed line) is clearly suboptimal. This is because it jointly updates the feature extractor and the linear separator, and the local updates diverge in different directions. On the contrary, FedLog clients learn universal local representations with fixed last linear layer, and the server is able to draw the linear separator as if it has seen all client data. Finally, the top and bottom right images show the training and testing accuracy v.s. local iterations. The testing data points are sampled i.i.d. from the same distribution. The results show: i) FedLog makes more progress in one global update round than other averaging based prior arts; ii) FedLog is resistant to over-fitting, as the difference between training and testing accuracy is small; iii) FedLog can learn universal local representations by fixing the last layer, even with different initialization and architectures of the feature extractor.

\subsection{Communication Cost}
\label{comm}
To show FedLog achieves better accuracy with less communication with non-i.i.d. clients, we conduct experiments on MNIST, CIFAR10, and CIFAR100. We compare FedLog and FedLog-C with the following baselines: i) FedAvg \citep{fedavg}, which averages the whole model each global update round; ii) LG-FedAvg \citep{lgfedavg}, which localizes bodies and averages heads. Two variants are reported: LG-FedAvg 1 which maintains one global layer and LG-FedAvg 2 which maintains two global layers; iii) FedPer \citep{fedper}, which localizes heads and averages bodies; iv) FedRep \citep{fedrep}, which also localizes heads and averages bodies, but trains local heads and bodies separately; v) CS-FL \citep{li2021communication}, a model compression technique that compresses messages with the compressed sensing framework; vi) FedBabu \citep{fedbabu}, which averages bodies and never updates heads; vii) FedProto \citep{fedproto}, which averages local feature representations by class and forces local models to learn similar representations; viii) FedDBE \citep{feddbe}, which averages both bodies and heads but accelerates convergence by learning a domain representation bias. Convolutional neural networks of the same architecture and initialization is used for all the algorithms. FedPer and FedRep localizes the last two layers.

We first distribute the training set into $(50, 100, 100)$ heterogeneous clients for (MNIST, CIFAR10, CIFAR100) respectively. Each client takes only $(2, 2, 10)$ classes. The testing set is distributed similarly to the clients, following the same distribution as the training data. Specially, for \textbf{MNIST only}, to simulate the difficult situation where clients do not have sufficient local data, we train on $5\%$ of the training set, while test on the whole testing set. All the clients start from the same initialization, and all the algorithms are run for $(100,100,150)$ global update rounds. The testing accuracy is recorded after each round. Hyperparameters are optimized beforehand through grid search. See Appendix D for the model architectures, hyperparameters used by each algorithm and other experiment details.

We report mean $\pm$ standard error of the testing accuracy resulting from $10$ seeds in Table~\ref{Tab:comm}. We measured the statistical significance of the results compared to FedLog-C with one-tailed Wilcoxon signed-rank tests \citep{wilcoxon}. Similarly, we report the total communication cost, namely the total size of messages transmitted between clients and the server, to reach a reasonable accuracy threshold (97\%, 83\%, and 53\% respectively). If it is never reached, we stop counting at the round with the highest accuracy. These thresholds are the largest integer of accuracy which all competitive algorithms (FedLog, LG-FedAvg, FedProto) have reached in 10 seeds. 

The results show FedLog can achieve statistically significant accuracy improvement compared to prior arts, with the least communication cost as small as $0.09\%$, $0.29\%$, and $3.44\%$ of the communication cost of FedAvg for MNIST, CIFAR10, and CIFAR100 respectively.  The closest baseline is FedProto, which has a slightly lower performance and requires more communication on CIFAR10 and CIFAR100 because it uses FedAvg instead of Bayesian inference for aggregation.

\begin{figure}[t]
  \centering\includegraphics[width=7cm,height=3.75cm]{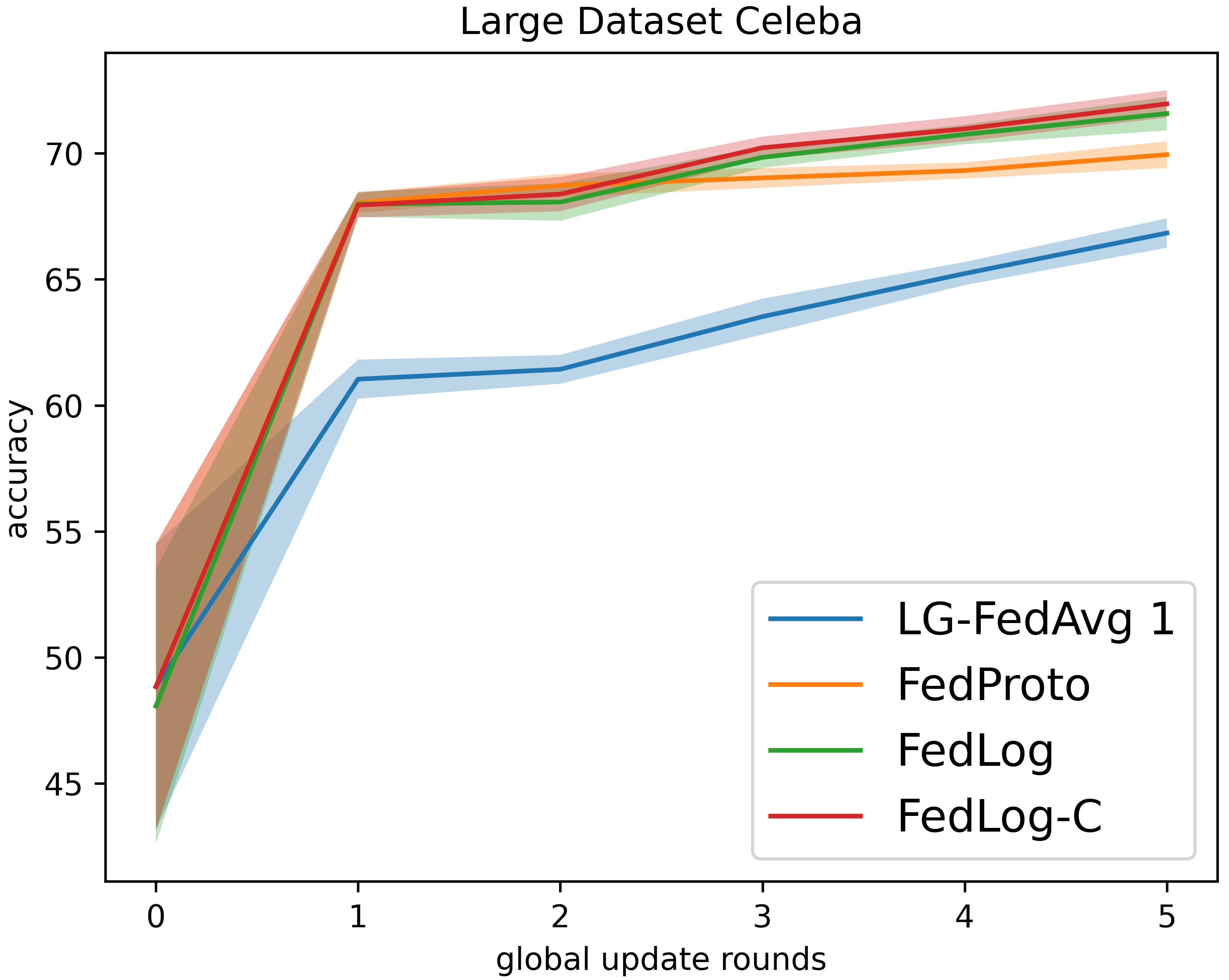}
  \includegraphics[width=0.4\textwidth]{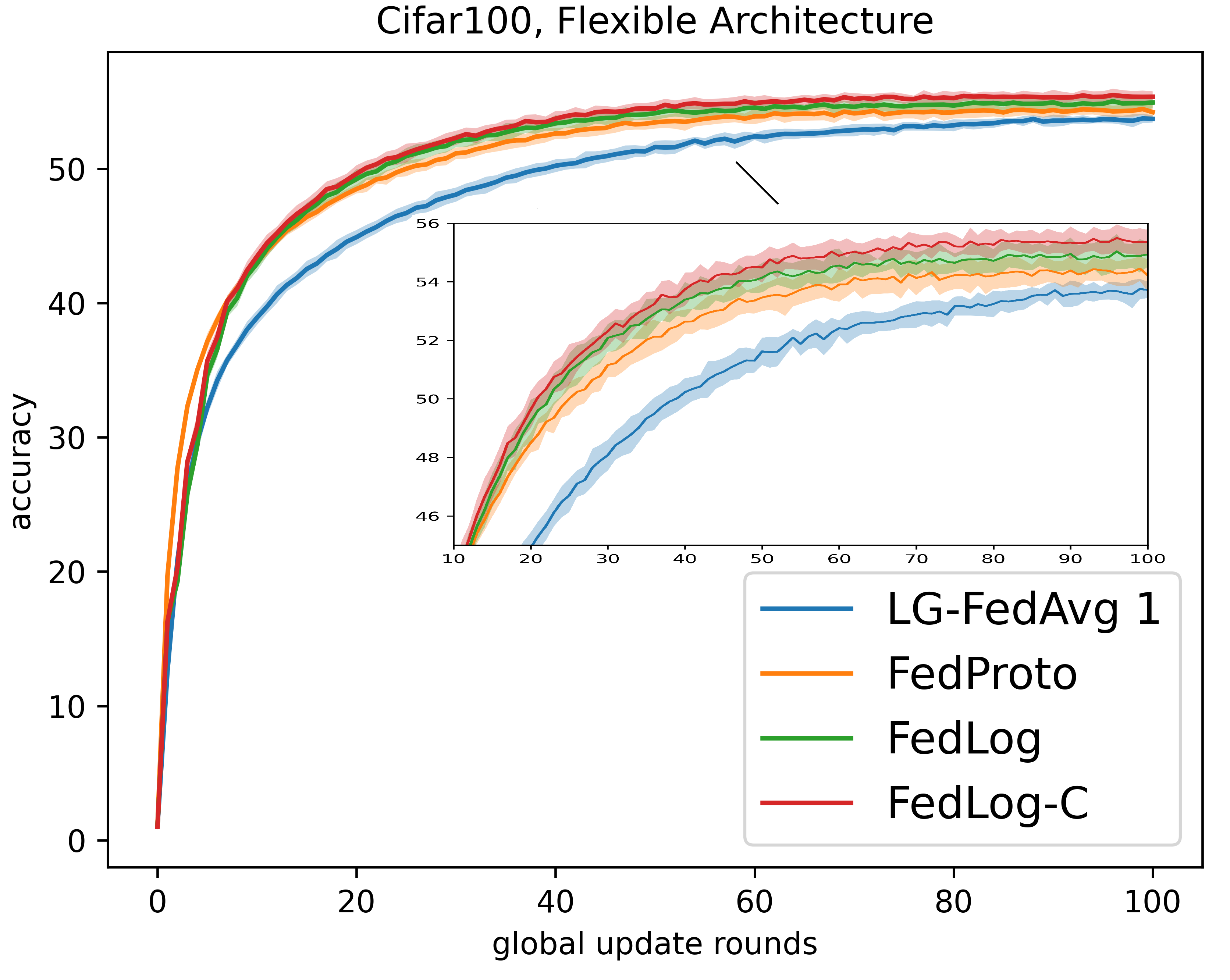}
  \includegraphics[width=7cm,height=3.75cm]{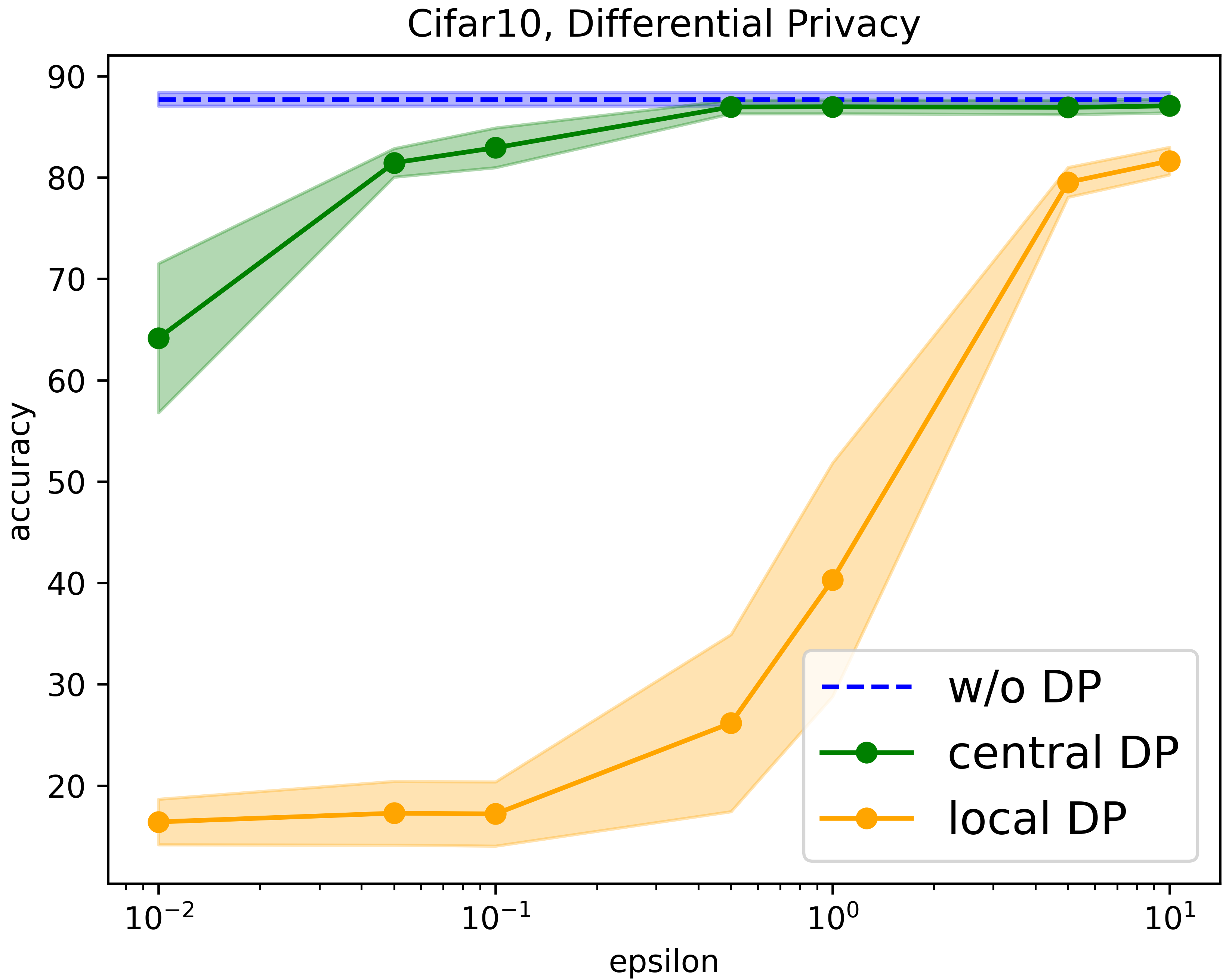}
  \caption{From top to bottom: Celeba, flexible architecture, and differential privacy results. Colored area shows mean $\pm$ standard deviation of the accuracy.}
  \label{fig:celeba}
\end{figure}

\subsection{Celeba}

We compare FedLog and FedLog-C to LG-FedAvg and FedProto on the large dataset Celeba \citep{celeba} preprocessed by LEAF \citep{caldas2018leaf}. We sampled 70838 images from 2360 clients. Each client represents a celebrity, and the local data are images of the same person. The task is to classify if the celebrity is smiling. MobileNetV2 \citep{sandler2018mobilenetv2} is implemented as the classifier model for all algorithms. To test how these algorithms perform under the situation where only a few communication rounds are available, we run every algorithm to optimum locally before global aggregation. The results are shown in the top graph of Fig. \ref{fig:celeba}. FedLog-C performs statistically significantly better than FedProto ({\bf 71.96$\pm$0.18} v.s. 69.98$\pm$0.16$^\Downarrow$, $p=0.001$), LG-FedAvg (66.84$\pm$0.19$^\Downarrow$, $p=0.001$), and FedLog (71.19$\pm$0.45$^\downarrow$, $p=0.05$).

\subsection{Flexible Architecture}

We simulate the situation where clients have different computational resources on CIFAR100. We randomly select half of the clients and assign them a smaller convolutional neural network, where the second last fully connected layer is removed. Different clients start from different local body initialization, but the global head is unified. We compare FedLog and FedLog-C to LG-FedAvg 1 and FedProto. As shown in the middle graph of Fig.~\ref{fig:celeba}, FedLog-C and FedLog converges faster than LG-FedAvg and FedProto. The accuracy of FedLog-C is also statistically significantly higher than FedProto ({\bf 55.86$\pm$0.11} v.s. 54.74$\pm$0.13$^\Downarrow$, $p = 0.001$), LG-FedAvg (54.03$\pm$0.08$^\Downarrow$, $p = 0.001$), and FedLog (55.31$\pm$0.15$^\Downarrow$, $p = 0.001$).

\subsection{Differential Privacy}

We conduct experiments to show the trade-off between privacy budget $\epsilon$ and the accuracy of FedLog on CIFAR10. We add an activation function to clip the extracted features to $b = 2$. Following a common practice in FL \cite{wei2020federated}, we set $\delta=0.01$. As shown in the bottom graph of Fig.~\ref{fig:celeba}, the accuracy of differentially private FedLog quickly grows back to optimum when $\epsilon \geq 0.5$, if the server is trusted or MPC is implemented so central DP is applicable. This is a strong privacy budget that shows FedLog performs well without sacrificing clients' privacy. Otherwise, local DP can have more impact on the model utility, but the accuracy is still acceptable when $\epsilon \geq 5.0$.

\section{Limitations}
One limitation of FedLog is that the algorithm works only for classification. The model is solely designed to mimic cross-entropy loss and therefore a different loss function with a different model would be needed for regression.  FedLog also assumes an exponential family distribution with a prior of the form $exp(-A(\boldsymbol{\eta}))$ and local transformed features distributed according to a mixture of Gaussians.  However the mixture of Gaussian assumption is mitigated in FedLog-C by introducing an auxiliary loss that helps satisfy the assumption.  A benefit of the exponential family and mixture of Gaussian assumptions is that the data summaries are provable sufficient statistics (Thm.~\ref{thm:sufficient-statistics}) and we obtain a closed form solution for the normalization constant (Eq.~\ref{eq:closed-form}). 

\section{Conclusion}
We proposed FedLog that shares local data summaries instead of model parameters. FedLog assumes an exponential family model on local representations, and learns a global linear separator with the summation of sufficient statistics. FedLog can learn universal local representations without sharing the bodies. Experiments show statistically significant improvements compared to prior arts, with the least communication cost. It is also effective with flexible architectures and formal DP guarantees. For future work, it would be interesting to generalize FedLog to regression.  % One limitation of FedLog is the assumption that the representations follow a mixture of Gaussians.  Relaxing this assumption in future work could further improve the performance of the approach.

\newpage
\section{Ackowledgment}
{Resources used in preparing this research at the University of Waterloo were provided by Huawei Canada, the province of Ontario and the government of Canada through CIFAR and companies sponsoring the Vector Institute.}
\bibliography{aaai25}

\clearpage

\appendix
\onecolumn

\renewcommand{\thetable}{A.\arabic{table}}
\renewcommand{\thetheorem}{A.\arabic{theorem}}
\setcounter{table}{0}

\section{A. Other Related Works}
\paragraph{Communication efficient FL.} Being orthogonal to FRL, communication efficient FL aims to directly decrease the communication overhead with optimization algorithms, client selection, and model compression \citep{wen2023survey}. First, since local training epochs affect the rounds of global communication needed \citep{fedavg}, researchers proposed different local optimization methods to reduce the communication rounds \citep{liu2020accelerating, wu2021fast, wu2022adaptive}. Second, some clients may contribute more to the global model, or are faster when uploading parameters. Thus, the global learning process can be accelerated by carefully selecting clients that meet these criteria \citep{liu2021joint,deng2021auction,lai2021oort,du2022bandwidth}. Additionally, the size of transmitted messages can be directly decreased by reducing or compressing the model parameters \cite{lu2020privacy,li2021cbfl,li2021communication}. However, most such algorithms are efficient at the cost of model accuracy \cite{cai2022aris}.
\paragraph{Bayesian FL.} Other Bayesian models have been explored to represent distributions over models and predictions in FL. The challenge is in the aggregation of the local posteriors. Various techniques have been proposed including personalized GPs~\citep{achituve2021personalized}, posterior averaging~\citep{al2020federated}, online Laplace approximation~\citep{liu2021bayesian}, Thompson sampling~\citep{dai2020federated}, MCMC sampling~\citep{pmlr-v151-vono22a}. These Bayesian FL techniques tend to emphasize calibration, approximating the posterior, or even different tasks. There is not much in common between them and our approach despite the use of Bayes theorem. 

\section{B. Sufficiency}
This section discusses formal definitions of sufficient and other statistics, based on Chapter 6 of \cite{casella2024statistical}.
\paragraph{Sufficient Statistics.}
\begin{definition}[6.2.1 in \cite{casella2024statistical}]
    A statistic ${\bf T}({\bf x})$ is a \textbf{sufficient} statistic for $\boldsymbol{\eta}$ if the conditional distribution of the sample ${\bf x}$ given the value of ${\bf T}({\bf x})$ does not depend on $\boldsymbol{\eta}$.
\end{definition}
Intuitively, a sufficient statistic captures all the information about $\boldsymbol{\eta}$ in ${\bf X}$. In the case of the exponential family, the following theorem applies:
\begin{theorem}[6.2.10 in \cite{casella2024statistical}]
    Let ${\bf x}_1, \cdots, {\bf x}_n$ be i.i.d. observations from an exponential family whose p.d.f. is given by
    $\Pr({\bf x}|\boldsymbol{\eta}) = h({\bf x})\exp(\boldsymbol{\eta}^\top {\bf T}({\bf x})-A(\boldsymbol{\eta}))$, then
    ${\bf T}({\bf x}_1, \cdots, {\bf x}_n) = \sum_{i=1}^{n} {\bf T}({\bf x}_i)$ is a sufficient statistic for $\boldsymbol{\eta}$.
    \label{thm:sufficient}
\end{theorem}

\paragraph{Minimal Sufficient Statistics.}
\begin{definition}[6.2.11 in \cite{casella2024statistical}]
    A sufficient statistic $\bf T(x)$ is \textbf{minimal} if, for any other sufficient statistic $\bf T'(x)$, $\exists$ a function $h$ such that $h({\bf T'(x)}) = {\bf T(x)}$. 
\end{definition}
A minimal sufficient statistic achieves the greatest possible data reduction for a sufficient statistic. Whether a sufficient statistic is minimal can be verified by the following theorem.
\begin{theorem}[6.2.13 in \cite{casella2024statistical}]
    If for all sample points $\bf x, x'$ from the distribution with p.d.f. $\Pr({\bf x}|\boldsymbol{\eta})$,
    $\frac{\Pr({\bf x}|\boldsymbol{\eta})}{\Pr({\bf x'}|\boldsymbol{\eta})}$ is independent of $\boldsymbol{\eta}$ iff $\bf T(x) = T(x')$,
    then $\bf T(x)$ is minimal sufficient.
    \label{thm:minimal}
\end{theorem}

\paragraph{Ancillary Statistics.}
\begin{definition}[6.2.16 in \cite{casella2024statistical}]
    A statistic $\bf S(x)$ is an \textbf{ancillary} statistic if it is independent of the parameters $\boldsymbol{\eta}$.
\end{definition}
As shown by the definition, an ancillary statistic on its own contains no information about the parameters $\boldsymbol{\eta}$.

\paragraph{Complete Statistics.}
\begin{definition}[6.2.21 in \cite{casella2024statistical}]
    A family of distributions is called \textbf{complete} if $\mathbb{E}_{\boldsymbol{\eta}}g({\bf T}) = 0, \forall \boldsymbol{\eta}$ implies $\Pr_{\boldsymbol{\eta}}(g({\bf T})=0)=1, \forall \boldsymbol{\eta}$. Then $\bf T(x)$ is a \textbf{complete} statistic.
\end{definition}
This definition is less intuitive and harder to interpret. We skip the details, but focus on the following theorems.
\begin{theorem}[6.2.25 in \cite{casella2024statistical}]
    Let ${\bf x}_1, \cdots, {\bf x}_n$ be i.i.d. observations from an exponential family whose p.d.f. is given by
    $\Pr({\bf x}|\boldsymbol{\eta}) = h({\bf x})\exp(\boldsymbol{\eta}^\top {\bf T}({\bf x})-A(\boldsymbol{\eta}))$, then
    ${\bf T}({\bf x}_1, \cdots, {\bf x}_n) = \sum_{i=1}^{n} {\bf T}({\bf x}_i)$ is a complete statistic for $\boldsymbol{\eta}$ if support of $\boldsymbol{\eta}$ is open.
    \label{thm:complete}
\end{theorem}
\begin{theorem}[Basu's Theorem, 6.2.24 in \cite{casella2024statistical}]
    If $\bf T(x)$ is a complete and minimal sufficient statistic, then $\bf T(x)$ is independent of every ancillary statistic.
    \label{thm:basu}
\end{theorem}
\paragraph{Proof of Our Claim}
We now prove our claim made in the main paper. We copy Theorem 2 from the main paper here.
\setcounter{theorem}{1}
\renewcommand{\thetheorem}{\arabic{theorem}}
\begin{theorem}
    If $({\boldsymbol{\phi}_1}, y_1), ({\boldsymbol{\phi}_2}, y_2), \cdots, ({\boldsymbol{\phi}_n}, y_n)$ are i.i.d. samples from the exponential family with ${\bf T}({\boldsymbol{\phi}}, y) = \boldsymbol{\phi}\otimes{\bf e}_y$, then ${\bf T}(({\boldsymbol{\phi}_1}, y_1), \cdots, ({\boldsymbol{\phi}_n}, y_n)) = \sum_{i=1}^n {\boldsymbol{\phi}_i}\otimes{\bf e}_{y_i}$ is a minimal sufficient statistic independent of every ancillary statistic. \label{thm:sufficient-statistics}
\end{theorem}
\begin{proof}
\begin{enumerate}
    \item ${\bf T}(({\boldsymbol{\phi}_1}, y_1), ({\boldsymbol{\phi}_2}, y_2), \cdots,({\boldsymbol{\phi}_n}, y_n))$ is sufficient by Thm. \ref{thm:sufficient}.
    \item ${\bf T}(({\boldsymbol{\phi}_1}, y_1), ({\boldsymbol{\phi}_2}, y_2), \cdots,({\boldsymbol{\phi}_n}, y_n))$ is complete by Thm. \ref{thm:complete}, since $\boldsymbol{\eta} \in \mathbb{R}^{m*n_{class}}$ is open.
    \item Let
\begin{align*}
    R &= \frac{\Pr(({\boldsymbol{\phi}_1}, y_1), ({\boldsymbol{\phi}_2}, y_2), \cdots, ({\boldsymbol{\phi}_n}, y_n)|\boldsymbol{\eta})}{\Pr(({\boldsymbol{\phi}'_1}, y'_1), ({\boldsymbol{\phi}'_2}, y’_2), \cdots, ({\boldsymbol{\phi}'_n}, y'_n)|\boldsymbol{\eta})} \\
    &= \exp(\boldsymbol{\eta}^\top(\sum_{i=1}^n {\boldsymbol{\phi}_i}\otimes{\bf e}_{y_i}-\sum_{i=1}^n {\boldsymbol{\phi}'_i}\otimes{\bf e}_{y'_i}))
\end{align*}
${\bf T}(({\boldsymbol{\phi}_1}, y_1), ({\boldsymbol{\phi}_2}, y_2), \cdots,({\boldsymbol{\phi}_n}, y_n))$ is also minimal by Thm. \ref{thm:minimal}, since $R$ is independent of $\boldsymbol{\eta}$ if and only if $${\bf T}(({\boldsymbol{\phi}_1}, y_1), \cdots, ({\boldsymbol{\phi}_n}, y_n)) = {\bf T}(({\boldsymbol{\phi}'_1}, y'_1), \cdots ({\boldsymbol{\phi}'_n}, y'_n))$$

\item ${\bf T}(({\boldsymbol{\phi}_1}, y_1), ({\boldsymbol{\phi}_2}, y_2), \cdots,({\boldsymbol{\phi}_n}, y_n))$ is independent of every ancillary statistic by Thm. \ref{thm:basu}.
\end{enumerate}
\end{proof}

\setcounter{theorem}{8}
\renewcommand{\thetheorem}{A.\arabic{theorem}}
\section{C. Differential Privacy}
It has been shown that FL algorithms that share large model parameters do not prevent privacy attacks through weight manipulation, GAN-based reconstruction, and large model memorization effects \citep{boenisch2021curious,mothukuri2021survey}. We argue that FedLog, which shares summations of sufficient statistics only, avoid these pitfalls closely related to model parameter sharing. Since ``addition'' is a non-invertible function, malicious attackers cannot recover features of individual data points. Since the local architecture and weights of the feature extractor is never shared in anyway, malicious attackers should not be able to reconstruct the original inputs, even if they are given the features of individual data points. We acknowledge that this argument is merely intuitive, and sharing data summaries could pose other risks of privacy leakage. To further guarantee users' privacy formally, we now extend FedLog to be differentially private.

($\epsilon, \delta$)-DP protects clients' privacy by adding noise to the shared information so that the adversaries cannot effectively tell if any record is included in the dataset (controlled by $\epsilon > 0$) at most times (controlled by $0 \leq \delta < 1$) \citep{dp}.
\begin{definition}
    A mechanism $M_{DP}$ satisfies $(\epsilon$, $\delta)$-DP if for any two datasets $\mathcal{D}, \mathcal{D}'$ that differ by only one record (i.e. $|(\mathcal{D}-\mathcal{D}') \cup (\mathcal{D}'-\mathcal{D})|=1$), and for any possible output $O \in Range(M_{DP})$,
    $$\Pr_{O \sim M_{DP}(\mathcal{D})}\left[\log(\frac{\Pr[M_{DP}(\mathcal{D})=O]}{\Pr[M_{DP}(\mathcal{D}')=O]}) > \epsilon \right] < \delta$$
\end{definition}
Intuitively, $(\epsilon, \delta)$-DP guarantees that the inner log ratio, considered as the information loss leaked to the adversaries, is bounded by the privacy budget $\epsilon$ with probability $\delta$. Usually, $\epsilon \leq 1$ is viewed as a strong protection, while $\epsilon \geq 10$ does not protect much. The magnitude of noise needed is usually determined by $\epsilon, \delta$ and the sensitivity of the function $f$, of which the results ($f(\mathcal{D})$ the revealed information) need protection. 
\begin{definition}
    The $L_p$ sensitivity of any function $f: \mathcal{D} \xrightarrow{} \mathbb{R}^n$ is $L_p(f) = \max_{\mathcal{D},\mathcal{D}'}||f(\mathcal{D})-f(\mathcal{D}')||_p$. $\mathcal{D}$ and $\mathcal{D}'$ differ by only one record.
\end{definition}

A commonly used mechanism is to add Gaussian noise to $f(\mathcal{D})$:
\begin{theorem}[\cite{kairouz2015composition}]
\label{thm:dp}
    For real-valued queries with sensitivity $L_2(f) > 0$, the mechanism that adds Gaussian noise with standard deviation $\sqrt{8k\ln(e+\epsilon/\delta)}L_2(f)/{\epsilon}$ satisfies $(\epsilon, \delta)$-differential privacy under $k$-fold adaptive composition, $\forall \epsilon > 0, \delta\in (0,1]$.
\end{theorem}

In FedLog, the only private information shared by clients is the summation of statistics ${\bf T}(\boldsymbol{\Phi}_c, {\bf y}_c)$, a vector of size $n_{class}*m$. Unfortunately, $L_2({\bf T})$ is unbounded for standard deep neural networks $\Tilde{\boldsymbol{\theta}}_c$, since the output features are usually unbounded. We need to clip the absolute values of the features to $b$, by simply adding an activation function to the last layer of the feature extractor
\begin{equation}
g(x) :=
    \begin{cases}
    b,& \text{if } x > b\\
    -b,& \text{if } x < -b\\
    x, & \text{otherwise}
\end{cases}
\end{equation}
We now prove Theorem 4 from the main paper.

\setcounter{theorem}{3}
\renewcommand{\thetheorem}{\arabic{theorem}}
\begin{theorem}
    If the absolute value of features are clipped to $b$ and there are in total $k$ global update rounds, FedLog messages satisfy $(\epsilon, \delta)$-DP with additive Gaussian noise ${\bf T}'(\boldsymbol{\Phi}_c, {\bf y}_c) := {\bf T}(\boldsymbol{\Phi}_c, {\bf y}_c) +\mathcal{N}({\bf 0}, \sigma^2{\bf I})$, where $\sigma = \sqrt{8k(1+(m-1)*b^2)\ln(e+\epsilon/\delta)}/{\epsilon}$.
\end{theorem}
\begin{proof}
We calculate the $L_2$ sensitivity as follows
\begin{align}
    \label{eq:2}
    L_2({\bf T}) &= \max_{\mathcal{D},\mathcal{D}'}||{\bf T}(\mathcal{D})-{\bf T}(\mathcal{D}')||_2\\
    \label{eq:3}
    &= \max_{\boldsymbol{\phi}, y}||{\bf T}(\boldsymbol{\phi}, y)||_2 \\
    \label{eq:4}
    &= ||[1, b, b, \cdots, b, 0, 0, \cdots, 0]||_2\\
    &= \sqrt{1+(m-1)*b^2}
\end{align}
Eq. \ref{eq:2} equals to Eq. \ref{eq:3} due to our definition of neighbouring datasets (adding or removing one record). Eq. \ref{eq:3} equals to Eq. \ref{eq:4} since one of our features is always 1, and there are at most $m-1$ other non-zero entries with maximum value of $b$.

Finally, we apply Thm.~\ref{thm:dp} to get $\sigma = \sqrt{8k(1+(m-1)*b^2)\ln(e+\epsilon/\delta)}/{\epsilon}$.
\end{proof}

\section{D. Experiment Details}

Communication cost, flexible architecture, and differential privacy experiments are run on 1 NVIDIA T4 GPU with 16GB RAM. Celeba experiments are run on 1 NVIDIA A40 GPU with 48GB RAM. Training data are normalized and randomly cropped and flipped. The architecture of CNNs used are listed in Table \ref{Tab:arc1}. Some important hyperparameters are listed in Table \ref{Tab:Hyp1}, \ref{Tab:Hyp2}, and \ref{Tab:Hyp3}. Most hyperparameters follow the experiment setting reported in LG-FedAvg. We make our code public in the supplementary materials, where further details can be found.

\section{E. Licences}

Yann LeCun and Corinna Cortes hold the copyright of MNIST dataset, which is a derivative work from original NIST datasets. MNIST dataset is made available under the terms of the Creative Commons Attribution-Share Alike 3.0 license.

The CIFAR-10 and CIFAR-100 are labeled subsets of the 80 million tiny images dataset. They were collected by Alex Krizhevsky, Vinod Nair, and Geoffrey Hinton, made public at \url{https://www.cs.toronto.edu/~kriz/cifar.html}{, the CIFAR homepage}.

The CelebA dataset is available for non-commercial research purposes only. See \url{https://mmlab.ie.cuhk.edu.hk/projects/CelebA.html}{, the CelebA homepage} for the full agreement.

\clearpage

\begin{table}[]
    \centering
    \begin{tabular}{c|c|c}
        \toprule
         MNIST & CIFAR10 & CIFAR100 \\
        \midrule
         nn.Conv2d(1, 10, kernel\_size=5) & nn.Conv2d(3, 6, kernel\_size=5) & nn.Conv2d(3, 6, kernel\_size=5) \\
         F.max\_pool2d(kernel\_size=2) & nn.MaxPool2d(2,2) & nn.MaxPool2d(2,2) \\
         nn.Conv2d(10, 20, kernel\_size=5) & nn.Conv2d(6, 16, kernel\_size=5) & nn.Conv2d(6, 16, kernel\_size=5) \\
         F.max\_pool2d(kernel\_size=2) & nn.MaxPool2d(2,2) &  nn.MaxPool2d(2,2) \\
         nn.Linear(320, 50) & nn.Linear(400, 120) & nn.Linear(400, 120)  \\
         nn.Linear(50, 10) & nn.Linear(120, 100) & nn.Linear(120, 100) \\
         & nn.Linear(100, 10) & nn.Linear(100, 100) \\
         \bottomrule
    \end{tabular}
    \caption{CNN architectures used in the communication cost experiment. Dropout layers and ReLu activation functions are omitted.}
    \label{Tab:arc1}
\end{table}

\begin{table}
  \centering
  \caption{Hyperparameters used in communication cost experiments for MNIST.}
  \begin{tabular}{c|c|c}
    \toprule
    Algorithm & Hyperparameter & Value\\
    \midrule
    \multirow{5}{*}{FedLog} & optimizer & Adam \\
    & body learning rate & 0.001 \\
    & head learning rate & 0.01 \\
    & batch size & 10 \\
    & local epochs & 5 \\
    \midrule
    \multirow{4}{*}{FedAvg} & optimizer & Adam \\
    & learning rate & 0.001 \\
    & batch size & 10 \\
    & local epochs & 5 \\
    \midrule
    \multirow{5}{*}{LG-FedAvg 1} & \# global layers & 1\\
    & optimizer & Adam \\
    & learning rate & 0.001 \\
    & batch size & 10 \\
    & local epochs & 5 \\
    \midrule
    \multirow{5}{*}{LG-FedAvg 2} & \# global layers & 2\\
    & optimizer & Adam \\
    & learning rate & 0.001 \\
    & batch size & 10 \\
    & local epochs & 5 \\
    \midrule
    \multirow{4}{*}{FedPer} & optimizer & Adam \\
    & learning rate & 0.001 \\
    & batch size & 10 \\
    & local epochs & 5 \\
    \midrule
    \multirow{5}{*}{FedRep} & optimizer & Adam \\
    & learning rate & 0.001 \\
    & batch size & 10 \\
    & body epochs & 5 \\
    & head epochs & 10 \\
    \midrule
    \multirow{7}{*}{CS-FL} & optimizer & Adam \\
    & phase 1 learning rate & 0.001 \\
    & phase 2 learning rate & 0.001 \\
    & sparcity & 0.005 \\
    & dimension reduction & 0.1 \\
    & batch size & 10 \\
    & local epochs & 5 \\
    \bottomrule
  \end{tabular}
\label{Tab:Hyp1}
\end{table}

\begin{table}
  \centering
  \caption{Hyperparameters used in communication cost experiments for CIFAR10.}
  \begin{tabular}{c|c|c}
    \toprule
    Algorithm & Hyperparameter & Value\\
    \midrule
    \multirow{5}{*}{FedLog} & optimizer & Adam \\
    & body learning rate & 0.0005 \\
    & head learning rate & 0.01 \\
    & batch size & 50 \\
    & local epochs & 1 \\
    \midrule
    \multirow{4}{*}{FedAvg} & optimizer & Adam \\
    & learning rate & 0.0005 \\
    & batch size & 50 \\
    & local epochs & 1 \\
    \midrule
    \multirow{5}{*}{LG-FedAvg 1} & \# global layers & 1\\
    & optimizer & Adam \\
    & learning rate & 0.0005 \\
    & batch size & 50 \\
    & local epochs & 1 \\
    \midrule
    \multirow{5}{*}{LG-FedAvg 2} & \# global layers & 2\\
    & optimizer & Adam \\
    & learning rate & 0.0005 \\
    & batch size & 50 \\
    & local epochs & 1 \\
    \midrule
    \multirow{4}{*}{FedPer} & optimizer & Adam \\
    & learning rate & 0.0005 \\
    & batch size & 50 \\
    & local epochs & 1 \\
    \midrule
    \multirow{5}{*}{FedRep} & optimizer & Adam \\
    & learning rate & 0.0005 \\
    & batch size & 50 \\
    & body epochs & 1 \\
    & head epochs & 10 \\
    \midrule
    \multirow{7}{*}{CS-FL} & optimizer & Adam \\
    & phase 1 learning rate & 0.001 \\
    & phase 2 learning rate & 0.01 \\
    & sparcity & 0.0005 \\
    & dimension reduction & 0.2 \\
    & batch size & 10 \\
    & local epochs & 1 \\
    \bottomrule
  \end{tabular}
\label{Tab:Hyp2}
\end{table}

\begin{table}
  \centering
  \caption{Hyperparameters used in communication cost experiments for CIFAR100.}
  \begin{tabular}{c|c|c}
    \toprule
    Algorithm & Hyperparameter & Value\\
    \midrule
    \multirow{5}{*}{FedLog} & optimizer & Adam \\
    & body learning rate & 0.0005 \\
    & head learning rate & 0.01 \\
    & batch size & 50 \\
    & local epochs & 3 \\
    \midrule
    \multirow{4}{*}{FedAvg} & optimizer & Adam \\
    & learning rate & 0.0005 \\
    & batch size & 50 \\
    & local epochs & 3 \\
    \midrule
    \multirow{5}{*}{LG-FedAvg 1} & \# global layers & 1\\
    & optimizer & Adam \\
    & learning rate & 0.0005 \\
    & batch size & 50 \\
    & local epochs & 3 \\
    \midrule
    \multirow{5}{*}{LG-FedAvg 2} & \# global layers & 2\\
    & optimizer & Adam \\
    & learning rate & 0.0005 \\
    & batch size & 50 \\
    & local epochs & 3 \\
    \midrule
    \multirow{4}{*}{FedPer} & optimizer & Adam \\
    & learning rate & 0.0005 \\
    & batch size & 50 \\
    & local epochs & 3 \\
    \midrule
    \multirow{5}{*}{FedRep} & optimizer & Adam \\
    & learning rate & 0.0005 \\
    & batch size & 50 \\
    & body epochs & 3 \\
    & head epochs & 3 \\
    \midrule
    \multirow{7}{*}{CS-FL} & optimizer & Adam \\
    & phase 1 learning rate & 0.001 \\
    & phase 2 learning rate & 0.01 \\
    & sparcity & 0.0005 \\
    & dimension reduction & 0.1 \\
    & batch size & 10 \\
    & local epochs & 1 \\
    \bottomrule
  \end{tabular}
\label{Tab:Hyp3}
\end{table}

\end{document}